\documentclass[11pt,letterpaper]{article}

\usepackage{coling2016}
\usepackage{comment}

\usepackage{times}
\usepackage{url}
\usepackage{latexsym}
\usepackage{amsmath}
\usepackage{graphicx}
\usepackage{color}
\usepackage{multirow}
\usepackage{amssymb}
\usepackage{float}
\usepackage{CJKutf8}

\usepackage{helvet}
\usepackage{bm}
\usepackage{amsbsy}

\newcommand*{\affaddr}[1]{#1} % No op here. Customize it for different styles.
\newcommand*{\affmark}[1][*]{\textsuperscript{#1}}
\newcommand*{\email}[1]{\texttt{#1}}
\newcommand{\kaishu}[1]{\begin{CJK*}{UTF8}{gkai}{#1}\end{CJK*}}

\title{ Chinese Poetry Generation with Planning based Neural Network  }

\author{%
	Zhe Wang\affmark[\dag], Wei He\affmark[\ddag], Hua Wu\affmark[\ddag], Haiyang Wu\affmark[\ddag], Wei Li\affmark[\ddag], Haifeng Wang\affmark[\ddag], Enhong Chen\affmark[\dag]\\	
	\affaddr{\affmark[\dag]University of Science and Technology of China, Hefei, China}\\
	\affaddr{\affmark[\ddag]Baidu Inc., Beijing, China}\\	
	\email{xiaose@mail.ustc.edu.cn,  cheneh@ustc.edu.cn}\\
	\email{\{hewei06, wu\_hua, wuhaiyang, liwei08, wanghaifeng\}@baidu.com}\\
}

\date{}
\begin{document}
	
	\maketitle
	
	\begin{abstract}
	
		Chinese poetry generation is a very challenging task in natural language processing. In this paper, we propose a novel two-stage poetry generating method which first plans the sub-topics of the poem according to the user's writing intent, and then generates each line of the poem sequentially, using a modified recurrent neural network encoder-decoder framework. The proposed planning-based method can ensure that the generated poem is coherent and semantically consistent with the user's intent. A comprehensive evaluation with human judgments demonstrates that our proposed approach outperforms the state-of-the-art poetry generating methods and the poem quality is somehow comparable to human poets.
		
	\end{abstract}
	
	\section{Introduction}
	
	\blfootnote{	
		%
		% for review submission
		%
		%\hspace{-0.65cm}  % space normally used by the marker
		%Place licence statement here for the camera-ready version, see
		%Section~\ref{licence} of the instructions for preparing a
		%manuscript.
		%
		% % final paper: en-uk version 
		%
		% \hspace{-0.65cm}  % space normally used by the marker
		% This work is licenced under a Creative Commons 
		% Attribution 4.0 International Licence.
		% Licence details:
		% \url{http://creativecommons.org/licenses/by/4.0/}
		% 
		% % final paper: en-us version 
		%
		\hspace{-0.65cm}  % space normally used by the marker
		This work is licensed under a Creative Commons 
		Attribution 4.0 International License.
		License details:
		\url{http://creativecommons.org/licenses/by/4.0/}
	}
	The classical Chinese poetry is a great and important heritage of Chinese culture. During the history of more than two thousand years, millions of beautiful poems are written to praise heroic characters,  beautiful scenery, love, friendship, etc. There are different kinds of Chinese classical poetry, such as Tang poetry and Song iambics. Each type of poetry has to follow some specific structural, rhythmical and tonal patterns. Table 1 shows an example of quatrain which was one of the most popular genres of poetry in China. The principles of a quatrain include: The poem consists of four lines and each line has five or seven characters; every character has a particular tone, Ping (the level tone) or Ze (the downward tone); the last character of the second and last line in a quatrain must belong to the same rhyme category \cite{wang2002summary}. With such strict restrictions, the well-written quatrain is full of rhythmic beauty.

	In recent years, the research of automatic poetry generation has received great attention. Most approaches employ rules or templates  \cite{tosa2008hitch,wu2009new,netzer2009gaiku,oliveira2009automatic,oliveira2012poetryme}, genetic algorithms  \cite{manurung2004evolutionary,zhou2010genetic,manurung2012using}, summarization methods \cite{yan2013poet} and statistical machine translation methods \cite{jiang2008generating,he2012generating} to generate poems. More recently, deep learning methods have emerged as a promising discipline, which considers the poetry generation as a sequence-to-sequence generation problem \cite{zhang2014chinese,Wang2016ChineseSI,yi2016generating}. These      methods usually generate the first line by selecting one line from the dataset of poems according to the user's writing intents (usually a set of keywords), and the other three lines are generated based on the first line and the previous lines. The user's writing intent can only affect the first line, and the rest three lines may have no association with the main topic of the poem, which may lead to semantic inconsistency when generating poems. In addition, topics of poems are usually represented by the words from the collected poems in the training corpus. But as we know, the words used in poems, especially poems written in ancient time, are different from modern languages. As a consequence, the existing methods may fail to generate meaningful poems if a user wants to write a poem for a modern term (e.g., Barack Obama).
	
	In this paper, we propose a novel poetry generating method which generates poems in a two-stage procedure: the contents of poems (``what to say") are first explicitly planned, and then surface realization (``how to say") is conducted. Given a user's writing intent which can be a set of keywords, a sentence or even a document described by natural language, the first step is to determine a sequence of sub-topics for the poem using a poem planning model, with each line represented by a sub-topic. The poem planning model decomposes the user's writing intent into a series of sub-topics, and each sub-topic is related to the main topic and represents an aspect of the writing intent. Then the poem is generated line by line, and each line is generated according to the corresponding sub-topic and the preceding generated lines, using a recurrent neural network based encoder-decoder model (RNN enc-dec). We modify the RNN enc-dec framework to support encoding of both sub-topics and the preceding lines. The planning based mechanism has two advantages compared to the previous methods. First, every line of the generated poem has a closer connection to user's writing intent. Second, the poem planning model can learn from extra knowledge source besides the poem data, such as large-scale web data or knowledge extracted from encyclopedias. As a consequence, it can bridge the modern concepts and the set of words covered by ancient poems. Take the term ``Barack Obama" as the example: using the knowledge from encyclopedias, the poem planning model can extend the user's query, Barack Obama, to a series of sub-topics such as outstanding, power, etc.,  therefore ensuring semantic consistency in the generated poems.
	
	The contribution of this paper is two-fold. First, we propose a planning-based poetry generating framework, which explicitly plans the sub-topic of each line. Second, we use a modified RNN encoder-decoder framework, which supports encoding of both sub-topics and the preceding lines, to generate the poem line by line.

	\begin{table}[t]
		\label{sec:example1}
		\small
		\centering
		\begin{tabular}{|cc|}
			
			\hline
			\begin{CJK*}{UTF8}{gkai}静夜思\end{CJK*} & Thoughts in a Still Night\\

			\begin{CJK*}{UTF8}{gkai}床前明月\textbf{光}，\end{CJK*} (P P Z Z P) & The luminous moonshine before my bed,\\

			\begin{CJK*}{UTF8}{gkai}疑是地上\textbf{霜}。\end{CJK*} (* Z Z P P) & Is thought to be the frost fallen on the ground.\\

			\begin{CJK*}{UTF8}{gkai}举头望明月，\end{CJK*} (* Z P P Z) & 	I lift my head to gaze at the cliff moon,\\

			\begin{CJK*}{UTF8}{gkai}低头思故\textbf{乡}。\end{CJK*} (P P Z Z P) & 	And then bow down to muse on my distant home.\\

			\hline
		\end{tabular}
		\caption{  An example of Tang poetry. The tone is shown at the end of each line. P represents the level-tone, and Z represents the downward-tone; * indicates that the tone can be either. The rhyming characters are in boldface.  }		
	\end{table}

	The rest of this paper is organized as follows. Section \ref{sec:related} describes some previous work on poetry generation  and compares our work with previous methods. Section \ref{sec:ppg} describes our planning based poetry generation framework. We introduce the datasets and experimental results in Section \ref{sec:experiments}. Section \ref{sec:conclusion} concludes the paper.

	\section{Related Work}
	\label{sec:related}
	
	Poetry generation is a challenging task in NLP.  Oliveira et al. \shortcite{oliveira2009automatic,oliveira2012poetryme,Oliveira2014AdaptingAG} proposed a poem generation method based on semantic and grammar templates.  \newcite{netzer2009gaiku} employed a method based on word association measures.  \newcite{tosa2008hitch} and \newcite{wu2009new} used a phrase search approach for Japanese poem generation. \newcite{Greene2010AutomaticAO} applied statistical methods to analyze, generate and translate rhythmic poetry. \newcite{Colton2012FullFACEPG} described a corpus-based poetry generation system that uses templates to construct poems according to the given constrains. \newcite{yan2013poet} considered the poetry generation as an optimization problem based on a summarization framework with several constraints. Manurung \shortcite{manurung2004evolutionary,manurung2012using} and \newcite{zhou2010genetic} used genetic algorithms for generating poems. An important approach to poem generation is based on statistical machine translation (SMT).  \newcite{jiang2008generating} used an SMT-based model in generating Chinese couplets which can be regarded as simplified regulated verses with only two lines. The first line is regarded as the source language and translated into the second line.  \newcite{he2012generating} extended this method to generate quatrains by translating the previous line to the next line sequentially. 
	
	Recently, deep learning methods achieve great success in poem generation. \newcite{zhang2014chinese} proposed a quatrain generation model based on recurrent neural network (RNN). The approach generates the first line from the given keywords with a recurrent neural network language model (RNNLM) \cite{mikolov2010recurrent} and then the subsequent lines are generated sequentially by accumulating the status of the lines that have been generated so far. \newcite{Wang2016ChineseSI} generated the Chinese Song iambics using an end-to-end neural machine translation model. The iambic is generated by translating the previous line into the next line sequentially. This procedure is similar to SMT, but the semantic relevance between sentences is better.  \newcite{Wang2016ChineseSI} did not consider the generation of the first line. Therefore, the first line is provided by users and must be a well-written sentence of the poem. \newcite{yi2016generating} extended this approach to generate Chinese quatrains. The problem of generating the first line is resolved by a separate neural machine translation (NMT) model which takes one keyword as input and translates it into the first line. \newcite{Marjan2016topical} proposed a poetry generation algorithm that first generates the rhyme words related to the given keyword and then generated the whole poem according to the rhyme words with an encoder-decoder model \cite{Sutskever2014}.
	
	Our work differs from the previous methods as follows. First, we don't constrain the user's input. It can be some keywords, phrases, sentences or even documents. The previous methods can only support some keywords or must provide the first line. Second, we use planning-based method to determine the topic of the poem according to the user's input, with each line having one specific sub-topic, which guarantees that the generated poem is coherent and well organized, therefore avoiding the problem of the previous method that only the first line is guaranteed to be related to the user's intent while the next lines may be irrelevant with the intention due to the coherent decay problem \cite{he2012generating,zhang2014chinese,Wang2016ChineseSI,yi2016generating}. Third, the rhythm or tone in 
	\cite{zhou2010genetic,yan2013poet,zhang2014chinese,yi2016generating,Marjan2016topical} is controlled by rules or extra structures, while our model can automatically learn constrains from the training corpus. Finally, our poem generation model has a simpler structure compared with those in \cite{zhang2014chinese,yi2016generating}.

	\section{Approaches}\label{sec:ppg}
	
	\begin{figure*}[t]
		\begin{center}
			\includegraphics[width=0.9\textwidth] {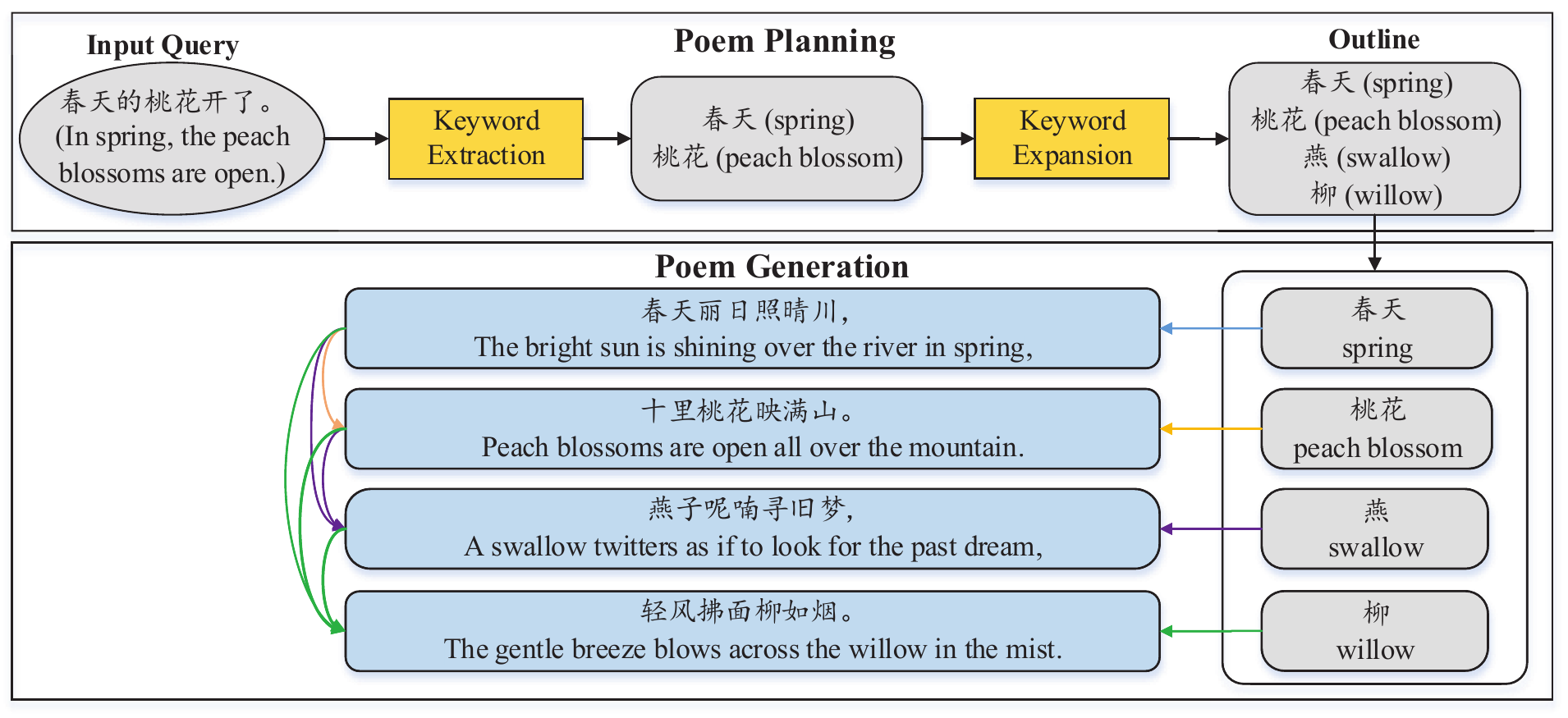} \\
			\caption{\label{fig1} Illustration of the planning based poetry generation framework. }
			
		\end{center}
	\end{figure*}	
	
	\subsection{Overview}
	
	Inspired by the observation that a human poet shall make an outline first before writing a poem, we propose a planning-based poetry generation approach (PPG) that first generates an outline according to the user's writing intent and then generates the poem. Our PPG system takes user's writing intent as input which can be a word, a sentence or a document, and then generates a poem in two stages: Poem Planning and Poem Generation. The two-stage procedure of PPG is illustrated in Figure \ref{fig1}.
	
	Suppose we are writing a poem that consists of $N$ lines with $l_i$ representing the $i$-th line of the poem. In the Poem Planning stage, the input query is transformed into $N$ keywords $(k_1,k_2,...,k_N)$, where $k_i$ is the $i$-th keyword that represents the sub-topic for the $i$-th line. In the Poem Generation stage, $l_i$ is generated by taking $k_i$ and $l_{1:i-1}$ as input, where $l_{1:i-1}$ is a sequence concatenated by all the lines generated previously, from $l_1$ to $l_{i-1}$. Then the poem can be generated sequentially, and each line is generated according to one sub-topic and all the preceding lines.
	
	\subsection{Poem Planning}
	\subsubsection{Keyword Extraction}	
	The user's input writing intent can be represented as a sequence of words. There is an assumption in the Poem Planning stage that the number of keywords extracted from the input query $Q$ must be equal to the number of lines $N$ in the poem, which can ensure each line takes just one keyword as the sub-topic. If the user's input query $Q$ is too long,  we need to extract the most important $N$ words and keep the original order as the keywords sequence to satisfy the requirement. 
	
	We use TextRank algorithm \cite{Mihalcea2004TextRankBO} to evaluate the importance of words.  It is a graph-based ranking algorithm based on PageRank \cite{Brin1998TheAO}.  Each candidate word is represented by a vertex in the graph and edges are added	between two words according to their co-occurrence; the edge weight is set according to the total count of co-occurrence strength of the two words. The TextRank score $S(V_i)$ is initialized to a default value (e.g. 1.0) and computed iteratively until convergence according to the following equation:
	\begin{equation}
	S({V_i}) = (1 - d) + d\sum\limits_{{V_j} \in E({V_i})} {\frac{{{w_{ji}}}}{{\sum\nolimits_{{V_k} \in E({V_j})} {{w_{jk}}} }}} S({V_j}),
	\end{equation}
	
	where $w_{ij}$ is the weight of the edge between node $V_j$ and $V_i$,  $E(V_i)$ is the set of vertices connected with $V_i$, and $d$ is a damping factor that usually set to 0.85 \cite{Brin1998TheAO}, and the initial score of $S(V_i)$ is set to 1.0.

	\subsubsection{Keyword Expansion}\label{sec:key_expansion}	
	If the user's input query $Q$ is too short to extract enough keywords, we need to expand some new keywords until the requirement of keywords number is satisfied. We use two different methods for keywords expansion.
	
	\textbf{RNNLM-based method.}
	We use a Recurrent Neural Network Language Model (RNNLM) \cite{mikolov2010recurrent} to predict the subsequent keywords according to the preceding sequence of keywords: $k_i=\arg \max_{k} P(k|k_{1:i-1})$, where $k_i$ is the $i$-th keyword and $k_{1:i-1}$ is the preceding keywords sequence.
	
	The training of RNNLM needs a training set consisting of keyword sequences extracted from poems, with one keyword representing the sub-topic of one line. We automatically generate the training corpus from the collected poems. Specifically, given a poem consisting of $N$ lines, we first rank the words in each line according to the TextRank scores computed on the poem corpus.
	%, where the TextRank score is computed on the poem corpus.
	Then the word with the highest TextRank score is selected as the keyword for the line. In this way, we can extract a keyword sequence for every poem, and generate a training corpus for the RNNLM based keywords predicting model. 
	
	%\subsection{Poem Planing for Out-domain Topic}
	\textbf{Knowledge-based method.}
	The above RNNLM-based method is only suitable for generating sub-topics for those covering by the collected poems. This method does not work when the user's query contains out-of-domain keywords, for example, a named entity not covered by the training corpus. 
	%The RNNLM-based keyword expansion, which is trained on the corpus of poems, is limited to the domain of topics. The method cannot work when the user's query contains out-domain keywords, for example, an named entity not covered by the corpus of poems. To expand from out-domain keywords, we propose a knowledge-based method that incorporate extra source of knowledge into the keyword expansion procedure.
	
	%As we known, most high-quality poems are written in the Tang and Song dynasties which were writen about 700 to 1,300 years ago. Almost all the topics of poems are within a closed domain that mostly contains common concepts in ancient times, such as sceneries, emotions, wars,  etc.. When a user wants to write a poem about a modern topic, the previous poem generating methods cannot solve this problem. In our planing based approach, we can plan appropriate sub-topics for a modern topic, then the poem generation model can generate a fluent but relative poem for the subject.
	
	To solve this problem, we propose a knowledge-based method that employs extra sources of knowledge to generate sub-topics. The extra knowledge sources can be used include encyclopedias, suggestions of search engines, lexical databases (e.g. WordNet), etc. Given a keyword $k_i$, the key idea of the method is to find some words that can best describe or interpret $k_i$. In this paper, we use the encyclopedia entries as the source of knowledge to expand new keywords from $k_i$. We retrieve those satisfying all the following conditions as candidate keywords: (1) the word is in the window of $[\textnormal{-}5,5]$ around $k_i$; (2) the part-of-speech of the word is adjective or noun; (3) the word is covered by the vocabulary of the poem corpus. Then the candidate words with the highest TextRank score are selected as the keywords.

	\subsection{Poem Generation}\label{sec:poem_generation}
	
	\begin{figure*}[t]
		\begin{center}
			\includegraphics[width=0.9\textwidth] {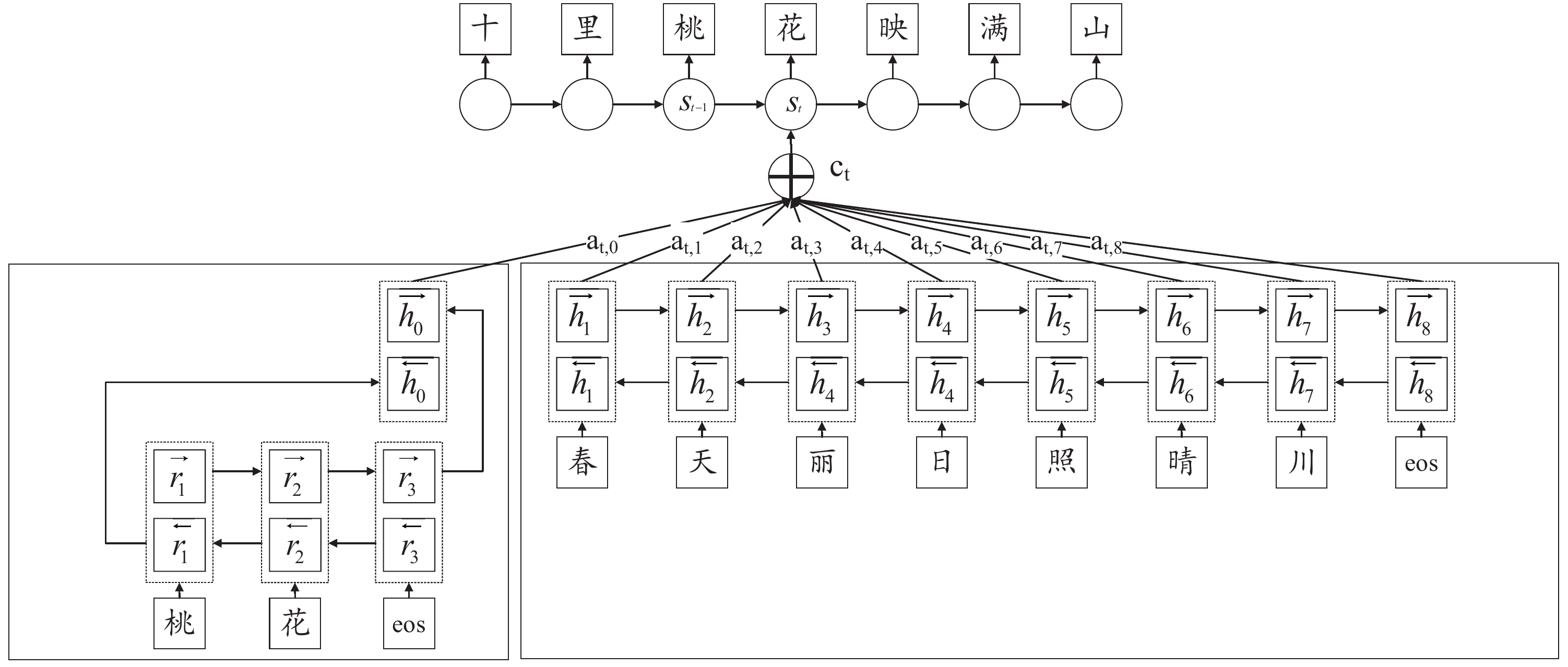} \\
			\caption{\label{fig2} An illustration of poem generation model. }
			
		\end{center}
	\end{figure*}    
	
	In the Poem Generation stage, the poem is generated line by line. Each line is generated by taking the keyword specified by the Poem Planning model and all the preceding text as input. This procedure can be considered as a sequence-to-sequence mapping problem with a slight difference that the input consists of two different kinds of sequences: the keyword specified by the Poem Planning model and the previously generated text of the poem. We modify the framework of an attention based RNN encoder-decoder (RNN enc-dec) \cite{bahdanau2014neural} to support multiple sequences as input.
	
	%We user all the preceding text as input to our model, which means the length of the input sequence is varied in [] 
	
	Given a keyword $\mathbf{k}$ which has $T_k$ characters, i.e. $\mathbf{k}=\{a_1,a_2,...,a_{T_k}\}$, and the preceding text $\mathbf{x}$ which has $T_x$ characters, i.e. $\mathbf{x}=\{x_1,x_2,...,x_{T_x}\}$, we first encode $\mathbf{k}$ into a sequence of hidden states $[r_1:r_{T_k}]$, and $\mathbf{x}$ into $[h_1:h_{T_x}]$, with bi-directional Gated Recurrent Unit (GRU) \cite{cho2014learning}  models. Then we integrate $[r_1:r_{T_k}]$ into a vector $r_c$ by concatenating the last forward state and the first backward state of $[r_1:r_{T_k}]$, where
	\begin{equation}
	r_{c}=\begin{bmatrix}
	\overrightarrow{r_{T_k}}\\
	\overleftarrow{r_{1}}
	\end{bmatrix}.
	\end{equation}
	
	We set $h_0=r_c$, then the sequence of vectors $\mathbf{h}=[h_0:h_{T_x}]$ represents the semantics of both $\mathbf{k}$ and $\mathbf{x}$, as illustrated in Figure \ref{fig2}. Notice that when we are generating the first line, the length of the preceding text is zero, i.e. $T_x=0$, then the vector sequence $\mathbf{h}$ only contains one vector, i.e. $\mathbf{h}=[h_0]$, therefore, the first line is actually generated from the first keyword.
	
	For the decoder, we use another GRU which maintains an internal status vector $s_t$, and for each generation step $t$, 
	the most probable output $y_t$ is generated based on $s_t$, context vector $c_t$ and previous generated output $y_{t-1}$. This can be formulated as follows:
	
	\begin{equation}
	y_{t} = \arg\max_y P(y|s_t,c_t,y_{t-1}).
	\end{equation}
	After each prediction, $s_t$ is updated by 
	\begin{equation}
	s_{t} = f(s_{t-1},c_{t-1},y_{t-1}).
	\end{equation}		
	$f(\cdot)$ is an activation function of GRU and $c_t$ is recomputed at each step by the alignment model:
	\begin{equation}
	{c_t} = \sum\limits_{j = 0}^{{T_h}-1} {{a_{tj}}{h_j}}.
	\end{equation}		
	$h_j$ is the $j$-th hidden state in the encoder's output. The weight $a_{tj}$ is computed by
	\begin{equation}
	{a_{tj}} = \frac{{\exp ({e_{tj}})}}{{\sum\nolimits_{k = 0}^{{T_h}-1} {\exp ({e_{tk}})} }},
	\end{equation}    
	where
	\begin{equation}
	{e_{tj}} = v_a^T\tanh ({W_a}{s_{t - 1}}+{U_a}{h_j}).
	\end{equation}		
	$e_{tj}$ is the attention score on $h_j$ at time step t. 		
	%With the decoder state $s_{t}$, the context $c_t$ and the last generated word $y_{t-1}$, 
	The probability of the next word $y_t$ can be defined as:
	\begin{equation}
	P(y_t|y_1,...,y_{t-1},\mathbf{x},\mathbf{k}) = g(s_t,y_{t-1},c_t),
	\end{equation}
	where $g(\cdot)$ is a nonlinear function that outputs the probability of $y_t$. 
	
	The parameters of the poem generation model are trained to maximize the log-likelihood of the training corpus:
	\begin{equation}
	\arg \max \sum_{n=1}^{N}log P(\mathbf{y_n}|\mathbf{x_n},\mathbf{k_n}).
	\end{equation}
	\begin{comment}
	\end{comment}

	\section{Experiments}\label{sec:experiments}
	
	\subsection{Dataset}

	In this paper, we focus on the generation of Chinese quatrain which has 4 lines and each line has the same length of 5 or 7 characters. We collected 76,859 quatrains from the Internet  and randomly chose 2,000 poems for validation, 2,000 poems for testing, and the rest for training.
	
	All the poems in the training set are first segmented into words using a CRF based word segmentation system. Then we calculate the TextRank score for every word. The word with the highest TextRank score is selected as the keyword for the line. In this way, we can extract a sequence of 4 keywords for every quatrain. From the training corpus of poems, we extracted 72,859 keyword sequences, which is used to train the RNN language model for keyword expansion (see section \ref{sec:key_expansion}). For knowledge-based expansion, we use Baidu Baike\footnote{A collaborative online encyclopedia provided by Chinese search engine Baidu: \url{http://baike.baidu.com}.}  and Wikipedia as the extra sources of knowledge.
	
	After extracting four keywords from the lines of a quatrain, we generate four triples composed of (the keyword, the preceding text, the current line), for every poem. Take the poem in Table \ref{sec:example1} as example,  the generated triples are shown in Table \ref{tbl:triples}. All the triples are used for training the RNN enc-dec model proposed in section \ref{sec:poem_generation}.
	
	\begin{table}[h]
		
		\small
		\centering
		\begin{tabular}{|c|c|c|}
			
			\hline
			Keyword & The Preceding Text & Current Line\\
			\hline
			\begin{CJK*}{UTF8}{gkai}床\end{CJK*} & $-$ & \begin{CJK*}{UTF8}{gkai}床前明月光\end{CJK*}\\
			
			\begin{CJK*}{UTF8}{gkai}霜\end{CJK*} & \begin{CJK*}{UTF8}{gkai}床前明月光\end{CJK*} & \begin{CJK*}{UTF8}{gkai}疑是地上霜\end{CJK*}\\
			
			\begin{CJK*}{UTF8}{gkai}明月\end{CJK*} & \begin{CJK*}{UTF8}{gkai}床前明月光; 疑是地上霜\end{CJK*} & \begin{CJK*}{UTF8}{gkai}举头望明月\end{CJK*}\\
			
			\begin{CJK*}{UTF8}{gkai}故乡\end{CJK*} & \begin{CJK*}{UTF8}{gkai}床前明月光; 疑是地上霜; 举头望明月\end{CJK*} & \begin{CJK*}{UTF8}{gkai}低头思故乡\end{CJK*}\\
			
			\hline
		\end{tabular}
		
		\caption{  Training triples extracted from the quatrain in Table \ref{sec:example1}. }	
		\label{tbl:triples}

	\end{table}

	\subsection{Training}
	\label{sec:train}
	
	For the proposed attention based RNN enc-dec model, we chose the 6,000 most frequently used characters as the vocabulary for both source and target sides. The word embedding dimensionality is 512 and initialized by word2vec \cite{mikolov2013efficient}. The recurrent hidden layers of the decoder and two encoders contained 512 hidden units. Parameters of our model were randomly initialized over a uniform distribution with support [-0.08,0.08]. The model was trained with the AdaDelta algorithm \cite{zeiler2012adadelta}, where the minibatch was set to be 128. The final model is selected according to the perplexity on the validation set.
	%We share the parametes of the two encoders ($enc_K$ and $enc_X$) to overcome the data sparsity. 
	
	\subsection{Evaluation}
	
	\subsubsection{Evaluation Metrics}
	It is well known that accurate evaluation of text generation system is difficult, such as the poetry generation and dialog response generation \cite{zhang2014chinese,schatzmann2005quantitative,seq2BF}. There are thousands of ways to generate an appropriate and relative poem or dialog response given a specific topic, the limited references are impossible to cover all the correct results. \newcite{liu2016not}  has recently shown that the overlap-based automatic evaluation metrics adapted for dialog responses, such as BLEU and METEOR, have little correlation with human evaluation. Therefore, we carry out a human study to evaluate the poem generation models. Following \cite{he2012generating,yan2013poet,zhang2014chinese}, we use four evaluation standards for human evaluators to judge the poems: ``Poeticness", ``Fluency", ``Coherence", ``Meaning". The detailed illustration can be seen in Table \ref{sec:human metrics}. The score of each aspect ranges from 1 to 5 with the higher score the better. Each system generates twenty 5-character quatrains and twenty 7-character quatrains. All the generated poems are evaluated by 5 experts and the rating scores are averaged as the final score.  
	
	\begin{table*}[t]
		\centering			
		\begin{tabular}{|c|c|}
			%	\hline \bf Metrics & \bf Descriptions  \\
			\hline
			Poeticness & Does the poem follow the rhyme and tone requirements ? \\
			\hline
			Fluency &Does the poem read smoothly and fluently? \\\hline
			Coherence & Is the poem coherent across lines? \\\hline
			Meaning & Does the poem have a certain meaning and artistic conception? \\
			
			\hline
		\end{tabular}
		\caption{\label{metrics} Evaluation standards in human judgement. }
		\label{sec:human metrics}
	\end{table*}

	\subsubsection{Baselines}    	
	
	We implemented several poetry generation methods as baselines and employed the same pre-processing method for all the methods.
	%conducted the same pre-generation process to all the systems for fairness. 

	\textbf{SMT}. A Chinese poetry generation method based on Statistical Machine Translation \cite{he2012generating}. A poem is generated iteratively by ``translating" the previous line into the next line.
	
	\textbf{RNNLM}. A method for generating textual sequences \cite{Graves2013}, which is proposed by \newcite{mikolov2010recurrent}. The lines of a poem are concatenated together as a character sequence which is used to train the RNNLM. 
	
	\textbf{RNNPG}. In the approach of RNN-based Poem Generator \cite{zhang2014chinese}, the first line is generated by a standard RNNLM and then all the other lines are generated iteratively based on a context vector encoded from the previous lines.
	
	\textbf{ANMT}. The Attention based Neural Machine Translation method. It considers the problem as a machine translation task, which is similar to the traditional SMT approach. The main difference is that in ANMT, the machine translation system is a standard  attention based RNN enc-dec framework \cite{bahdanau2014neural}.

	\begin{table*}[t]
		\centering
		\resizebox{\textwidth}{!}
		{
			%\begin{tabular}{|l|c|c|c|c|c|c|c|c|c|c| }
			\begin{tabular}{|l|l|l|l|l|l|l|l|l|l|l| }
				\hline
				\multicolumn{1}{|l|}{\multirow{2}[0]{*}{\bf Models}} & \multicolumn{2}{|c|}{\bf Poeticness} & \multicolumn{2}{c|}{\bf Fluency} & \multicolumn{2}{c|}{\bf Coherence} & \multicolumn{2}{c|}{\bf Meaning} & \multicolumn{2}{c|}{\bf Average} \\
				\cline{2-11}
				\multicolumn{1}{|l|}{} & \bf 5-char &\bf  7-char &\bf  5-char &\bf  7-char &\bf  5-char &\bf  7-char &\bf  5-char &\bf  7-char &\bf  5-char &\bf  7-char \\\hline
				SMT   & 3.25  & 3.22  & 2.81  & 2.48  & 3.01  & 3.16  & 2.78  & 2.45  & 2.96  & 2.83  \\
				RNNLM   & 2.67  & 2.55  & 3.13  & 3.42  & 3.21  & 3.44  & 2.90  & 3.08  & 2.98  & 3.12  \\
				RNNPG & 3.85  & 3.52  & 3.61  & 3.02  & 3.43  & 3.25  & 3.22  & 2.68  & 3.53  & 3.12  \\
				%ANMT  & \textbf{4.39}  & 4.11  & 4.52  & 4.49  & 4.16  & 4.24  & 4.05  & 4.24  & 4.28  & 4.27  \\
				%PPG   & 4.23  & \textbf{4.26}  & \textbf{4.70}  & \textbf{4.75}  & \textbf{4.49}  & \textbf{4.67}  & \textbf{4.54}  & \textbf{4.59}  & \textbf{4.49}  & \textbf{4.57}  \\
				%\hline
				%HUMAN & 4.26 & 4.05 & 4.60 & 4.36 & 4.35 & 4.40 & 4.43 & 4.44 & 4.41 & 4.31 \\
				ANMT  & \textbf{4.34}  & 4.04  & \textbf{4.61}  & 4.45    & 4.05  & 4.01  & $4.09$  & 4.04  & 4.27 & 4.14 \\
				PPG   & 4.11  & \textbf{4.15}  & 4.58  & $\textbf{4.56}^{\textbf{*}}$  & $\textbf{4.29}^{\textbf{*}}$  & $\textbf{4.49}^{\textbf{**}}$ & $\textbf{4.46}^{\textbf{**}}$  & $\textbf{4.51}^{\textbf{**}}$  & $\textbf{4.36}^{\textbf{**}}$  & $\textbf{4.43}^{\textbf{**}}$  \\
				%HUMAN & 4.34 & 4.28 & 4.49 & 4.63 & 4.45 & 4.68 & 4.48 & 4.68 & 4.44 & 4.57 \\
				\hline
				
			\end{tabular}
		}
		\caption{%\label{font-table} 
			Human evaluation results  of all the systems. Diacritics $^{**}$ (p \textless 0.01) and $^*$ (p \textless 0.05) indicate that our model (PPG) is significantly better than all other systems. }
		\label{sec:human}
		
	\end{table*}
	
	\subsubsection{Results}\label{sec:result}
	
	The results of the human evaluation are shown in Table \ref{sec:human}. We can see that our proposed method, Planning based Poetry Generation (PPG), outperforms all baseline models in average scores. The results are consistent with both settings of 5-character and 7-character poem generations.
	
	The poems generated by SMT are better in Poeticness than RNNLM, which demonstrates that the translation based method can better capture the mapping relation between two adjacent lines. ANMT is a strong baseline which performs better than SMT, RNNLM and RNNPG, but lower than our approach. Both ANMT and PPG use the attention based enc-dec framework. The main difference is that our method defines the sub-topics for each line before generating the poem. The ANMT method just translates the preceding text into the next line. Without the guide of sub-topics, the system tends to generate more general but less meaningful results. In contrast, our approach explicitly considers the keywords, which has better controls of the sub-topic for every line. From the results of the human evaluation, it can be seen that the proposed method obtained very close performances in Poeticness and Fluency compared with ANMT but much higher Coherence and Meaning scores, which verified the effectiveness of the sub-topic prediction model.

	%PPG (full) and PPG (shared encoders) are all outperforms the PPG (no outline) and RNNLM as they have the ability to take advantage from the outline which is helpful in the generation process.	We can also see that the PPG (full) does not show significant benefit (only 2.67 perplexity improvement) compared to PPG (shared encoders) . This is because although the effects of the outline and previously lines on the generation process are different, the difference is not particularly large. Another reason might be that the encoders with shared parameters are strong enough to learn the differences well.	PPG (no outline) has better performances than RNNLM which is not surprising, as the attention structure has more advantage in modeling long-range dependents. 

	%\subsection{Discussions}
	
	\subsection{Automatic Generation vs. Human Poet}

	We conducted an interesting evaluation that directly compares our automatic poem generation system with human poets, which is similar to the Turing Test \cite{turing1950computing}. We randomly selected twenty poems from the test set, which are written by ancient Chinese poets. We used the titles of these poems as the input and generated 20 poems by our automatic generation system. Therefore, the machine-generated poems were under the same subject with human-written poems. Then we asked some human evaluators to distinguish the human-written poems from machine-generated ones. We had 40 evaluators in total. All of them were well-educated and had Bachelor or higher degree. Four of them were professional in Chinese literature and were assigned to the Expert Group. The other thirty-six evaluators were assigned to the Normal Group. In the blind test, we showed a pair of poems and their title to the evaluator at each time, and the evaluator was asked to choose from three options: (1) poem A is written by the human; (2) poem B is written by the human; (3) cannot distinguish which one is written by the human.
	%If the evaluator cannot make the discision, he can give up this pair of poems.

	The evaluation results are shown in Table \ref{sec:turing}. We can see that 49.9\% of the machine-generated poems are wrongly identified as the human-written poems or cannot be distinguished by the normal evaluators. But for expert evaluators, this number drops to 16.3\%. We can draw two conclusions from the result: (1) under the standard of normal users, the quality of our machine-generated poems is very close to human poets; (2) but from the view of professional experts, the machine-generated poems still have some obvious shortages comparing to human-written poems.	Table \ref{sec:example3} gives an example for a pair of poems selected from our blind test. 
	
	\begin{table}[t]			
		\centering
		\resizebox{\textwidth}{!}{
			\begin{tabular}{|c|c|c|c|c|}
				\hline
				& Wrongly Identified MP as HP & Cannot Distinguish & Successfully Identified HP as HP\\
				\hline
				Normal Group &  38.6\% & 11.3\% &50.1\%  \\
				Expert Group &  6.3\% & 10.0\% & 83.7\% \\
				\hline
				
			\end{tabular}%
		}
		\caption{\label{font-table} Blind test to distinguish Human-written Poems (HP) from Machine-generated Poems (MP). }
		\label{sec:turing}			
	\end{table}

	\begin{table}[t]    
		\centering
		\resizebox{\textwidth}{!}{
			\begin{tabular}{|c|c|}
				
				\hline
				\begin{CJK*}{UTF8}{gkai}秋夕湖上\end{CJK*}& \begin{CJK*}{UTF8}{gkai}秋夕湖上\end{CJK*}\\
				By a Lake at Autumn Sunset& By a Lake at Autumn Sunset\\
				\begin{CJK*}{UTF8}{gkai}一夜秋凉雨湿衣，\end{CJK*}& \begin{CJK*}{UTF8}{gkai}荻花风里桂花浮，\end{CJK*} \\
				A cold autumn rain wetted my clothes last night, & The wind blows reeds with osmanthus flying, \\
				\begin{CJK*}{UTF8}{gkai}西窗独坐对夕晖。\end{CJK*}&\begin{CJK*}{UTF8}{gkai}恨竹生云翠欲流。\end{CJK*} \\
				And I sit alone by the window and enjoy the sunset.  & And the bamboos under clouds are so green as if to flow down.\\
				\begin{CJK*}{UTF8}{gkai}湖波荡漾千山色，\end{CJK*}& \begin{CJK*}{UTF8}{gkai}谁拂半湖新镜面，\end{CJK*}\\
				With mountain scenery mirrored on the rippling lake, & The misty rain ripples the smooth surface of lake,\\
				\begin{CJK*}{UTF8}{gkai}山鸟徘徊万籁微。\end{CJK*} & \begin{CJK*}{UTF8}{gkai}飞来烟雨暮天愁。\end{CJK*}\\
				A silence prevails over all except the hovering birds.  &  And I feel blue at sunset .   \\
				
				\hline	
			\end{tabular}
		}
		\caption{\label{font-table} A pair of poems selected from the blind test. The left one is a machine-generated poem, and the right one is written by Shaoti Ge, a poet lived in the Song Dynasty.} 
		%The two poems are selected from our turing-like test. The left poem is generated by our PPG system. The right poem was written by an ancient poet in Song dynasty.}
		
		\label{sec:example3}    
	\end{table}

	\begin{table}[t]    
		\centering
		\resizebox{\textwidth}{!}{
			\begin{tabular}{|c|c|}
				
				\hline
				\begin{CJK*}{UTF8}{gkai}啤酒\end{CJK*} & \begin{CJK*}{UTF8}{gkai}冰心\end{CJK*} \\
				Beer& Xin Bing\\
				\begin{CJK*}{UTF8}{gkai}今宵啤酒两三缸，\end{CJK*}& \begin{CJK*}{UTF8}{gkai}一片冰心向月明，\end{CJK*} \\
				I drink glasses of beer tonight, & I open up my pure heart  to the moon,   \\
				\begin{CJK*}{UTF8}{gkai}杯底香醇琥珀光。\end{CJK*}&\begin{CJK*}{UTF8}{gkai}千山春水共潮生。\end{CJK*} \\
				With the bottom of the glass full of aroma and amber light.   & With the spring river flowing past mountains.\\
				\begin{CJK*}{UTF8}{gkai}清爽金风凉透骨，\end{CJK*}& \begin{CJK*}{UTF8}{gkai}繁星闪烁天涯路，\end{CJK*}\\
				Feeling cold as the autumn wind blows, & Although my future is illuminated  by stars,\\
				\begin{CJK*}{UTF8}{gkai}醉看明月挂西窗。\end{CJK*} & \begin{CJK*}{UTF8}{gkai}往事萦怀梦里行。\end{CJK*}\\
				I get drunk and enjoy the moon in sight by the west window. & The past still lingers in my dream.   \\
				
				\hline	
			\end{tabular}
		}
		
		\caption{\label{font-table} Examples of poems generated from titles of modern concepts. }
		\label{sec:example4}
	\end{table}

	\subsection{Generation Examples}
	Besides the ancient poems in Table \ref{sec:example3}, our method can generate poems based any modern terms. Table \ref{sec:example4} shows some examples. The title of the left poem in Table \ref{sec:example4} is \emph{\kaishu{啤酒 (beer)}}, the keywords given by our poem planning model are \emph{\kaishu{啤酒 (beer)}}, \emph{\kaishu{香醇 (aroma)}}, \emph{\kaishu{清爽 (cool)}} and \emph{\kaishu{醉 (drunk)}}. The title of the right one is a named entity \emph{\kaishu{冰心 (Xin Bing)}}, who was a famous writer. The poem planning system generates three keywords besides \emph{\kaishu{冰心 (Xin Bing)}}: \emph{\kaishu{春水 (spring river)}}, \emph{\kaishu{繁星 (stars)}} and \emph{\kaishu{往事 (the past)}}, which are all related to the writer's works.

	\section{Conclusion and Future Work}\label{sec:conclusion}
	In this paper, we proposed a novel two-stage poetry generation method which first explicitly decomposes the user's writing intent into a series of sub-topics, and then generates a poem iteratively using a modified attention based RNN encoder-decoder framework. The modified RNN enc-dec model has two encoders that can encode both the sub-topic and the preceding text. The evaluation by human experts shows that our approach outperforms all the baseline models and the poem quality is somehow comparable to human poets. We have also demonstrated that using encyclopedias as an extra source of knowledge, our approach can expand users' input into appropriate sub-topics for poem generation. In the future, we will investigate more methods for topic planning, such as PLSA, LDA or word2vec. We will also apply our approach to other forms of literary genres e.g. Song iambics, Yuan Qu etc., or poems in other languages.
	
	\section{Acknowledgments}\label{sec:akg}
	This research was supported by the National Basic Research Program of China (973 program No. 2014CB340505), the National Key Research and Development Program of China (Grant No. 2016YFB1000904), the National Science Foundation for Distinguished Young Scholars of China (Grant No. 61325010) and the Fundamental Research Funds for the Central Universities of China (Grant No. WK2350000001). We would like to thank Xuan Liu, Qi Liu, Tong Xu, Linli Xu,  Biao Chang and the anonymous reviewers for their insightful comments and suggestions. 
	
	\label{sec:length}
	
	\label{sec:blind}

	\bibliography{emnlp2016}
	\bibliographystyle{emnlp2016}

\end{document}